\documentclass[10pt,twocolumn,letterpaper]{article}

\usepackage{cvpr}              

\usepackage{graphicx}
\usepackage{amsmath}
\usepackage{amssymb}
\usepackage{booktabs}

\usepackage{multirow}
\usepackage{xcolor}
\usepackage{colortbl}

\usepackage[pagebackref,breaklinks,colorlinks]{hyperref}

\usepackage[capitalize]{cleveref}
\crefname{section}{Sec.}{Secs.}
\Crefname{section}{Section}{Sections}
\Crefname{table}{Table}{Tables}
\crefname{table}{Tab.}{Tabs.}

\definecolor{mygray}{gray}{0.9}

\def\cvprPaperID{10178} 
\def\confName{CVPR}
\def\confYear{2023}

\begin{document}

\title{Privacy-preserving Adversarial Facial Features}

\author{Zhibo Wang$^{\dagger, \wr, \ast}$, He Wang$^{\dagger}$, Shuaifan Jin$^{\dagger}$, Wenwen Zhang$^{\ddagger}$, Jiahui Hu$^{\dagger}$, Yan Wang$^{\ddagger}$ \\ Peng Sun$^{\sharp}$, Wei Yuan$^{\natural}$, Kaixin Liu$^{\natural}$, Kui Ren$^{\dagger}$ \\
$^{\dagger}$School of Cyber Science and Technology, Zhejiang University, P. R. China \\
$^{\wr}$ZJU-Hangzhou Global Scientific and Technological Innovation Center \quad $^{\ddagger}$Alibaba Group, P. R. China\\
$^{\sharp}$College of Computer Science and Electronic Engineering, Hunan University, P. R. China \\ 
$^{\natural}$School of Cyber Science and Engineering, Wuhan University, P. R. China\\
{\tt\small \{zhibowang, wanghe\_71, shuaifanjin\}@zju.edu.cn, karida.zww@alibaba-inc.com, jiahuihu@zju.edu.cn}\\
{\tt\small wy84378@alibaba-inc.com, psun@hnu.edu.cn, \{wyuan, kxliu777\}@whu.edu.cn, kuiren@zju.edu.cn}
}
\maketitle
\newcommand\blfootnote[1]{%
\begingroup
\renewcommand\thefootnote{}\footnote{#1}%
\addtocounter{footnote}{-1}%
\endgroup
}
\blfootnote{This manuscript was accepted by CVPR 2023.}

\maketitle

\begin{abstract}
\vspace{-3mm}

Face recognition service providers protect face privacy by extracting compact and discriminative facial features (representations) from images, and storing the facial features for real-time recognition. However, such features can still be exploited to recover the appearance of the original face by building a reconstruction network. Although several privacy-preserving methods have been proposed, the enhancement of face privacy protection is at the expense of accuracy degradation. In this paper, we propose an adversarial features-based face privacy protection (AdvFace) approach to generate privacy-preserving adversarial features, which can disrupt the mapping from adversarial features to facial images to defend against reconstruction attacks. To this end, we design a shadow model which simulates the attackers’ behavior to capture the mapping function from facial features to images and generate adversarial latent noise to disrupt the mapping. The adversarial features rather than the original features are stored in the server's database to prevent leaked features from exposing facial information. Moreover, the AdvFace requires no changes to the face recognition network and can be implemented as a privacy-enhancing plugin in deployed face recognition systems. Extensive experimental results demonstrate that AdvFace outperforms the state-of-the-art face privacy-preserving methods in defending against reconstruction attacks while maintaining face recognition accuracy.

\end{abstract}


\vspace{-4mm}
\section{Introduction}
\label{sec:intro}
\vspace{-2mm}
Face recognition is a way of identifying an individual's identity using their face, which has been widely used in many security-sensitive applications. Undoubtedly, biometric facial images are private and discriminative information to each person that should be protected. Recently, much attention has been paid to privacy protection, such as the General Data Protection Regulation, making the preservation of face privacy increasingly important. In order to avoid direct leakage of facial images, mainstream face recognition systems usually adopt a client-server mode that extracts features from facial images with a feature extractor on the client side and stores the facial features rather than facial images on the server side for future online identification. As facial features suppress the visual information of faces, face privacy protection can be realized to some extent. 

However, recent studies showed that it is possible to reconstruct original images from facial features, which is called reconstruction attack, 
including optimization-based \cite{fredrikson2015model, razzhigaev2020black} and learning-based reconstruction attacks \cite{dosovitskiy2016inverting, zhmoginov2016inverting, mai2018reconstruction, he2019model}. The former gradually adjusts the pixels of the input image to make the output of the feature extractor as close as possible to a particular feature until the facial image (the input image) corresponding to this feature is reconstructed \cite{fredrikson2015model, razzhigaev2020black}. The latter trains a feature-image decoder with a de-convolutional neural network (D-CNN) to reconstruct images directly from facial features \cite{dosovitskiy2016inverting, zhmoginov2016inverting, mai2018reconstruction, he2019model}. These studies imply that existing face recognition systems suffer from severe privacy threats once the features in their database were leaked. Therefore, it is essential to provide approaches to prevent facial features from being reconstructed.

Several approaches have been proposed to protect face privacy. \cite{gentry2011implementing, mai2020secureface, kou2021efficient, abdalla2015simple} transform the features into the encrypted space and perform face recognition based on the cryptographic primitives and security protocols, which however bear prohibitive computation and communication costs for face recognition systems. \cite{chamikara2020privacy, Mao2018APD} utilize differential privacy to protect face privacy by perturbing features with noises, which however suffers from a significant accuracy drop in face recognition. \cite{li2021deepobfuscator, xiao2020adversarial} proposed adversarial training-based methods that retrain the main task network (e.g., gender classification from facial images) using adversarial training between the reconstruction network and the main task network to generate the privacy-preserving features directly. However, \cite{li2021deepobfuscator} demonstrated that facial features learned from adversarial training significantly compromise accuracy when dealing with face recognition tasks. Recently, several frequency domain-based methods \cite{ji2022privacy, mi2022duetface} were proposed, which transform raw images into the frequency domain and remove features' critical channels used for visualization to protect face privacy. However, \cite{ji2022privacy} struggles with the trade-off between accuracy and privacy protection and our experimental results demonstrate that \cite{mi2022duetface} actually cannot resist powerful reconstruction attacks. In addition, both the adversarial learning-based and the frequency domain-based methods require retraining the face recognition network, which is not applicable to deployed face recognition systems.


In this paper, we aim to propose a novel approach to generate privacy-preserving facial features which are able to thwart reconstruction attacks as well as maintain satisfactory recognition accuracy. Undoubtedly, it is non-trivial to realize this objective. The first challenge is \textit{how to defend against reconstruction attacks under the black-box setting}. An attacker may utilize different reconstruction networks, which are unknown to the face recognition systems in advance, to reconstruct images from facial features. How to enable the generated facial features to defend against such unknown and different reconstruction attacks is very challenging. The second challenge is \textit{how to disrupt the visual information embedded in facial features while keeping the recognition accuracy}. Since visual information is somewhat critical to face recognition, disrupting visual information may incur a reduction in recognition accuracy. The last challenge is \textit{how to generate privacy-preserving features without changing the face recognition network}. Once a face recognition network is deployed, it would be expensive to retrain the network and redeploy it to millions of clients. Therefore, a plug-in module is more welcomed for the deployed face recognition systems.

To address the above challenges, we propose an \emph{adversarial features-based face privacy protection (AdvFace)} method, which generates the privacy-preserving adversarial features against reconstruction attacks. The intuition of AdvFace is to disrupt the mapping from features to facial images by obfuscating features with adversarial latent noise to maximize the difference between the original images and the reconstructed images from the features. To this end, we train a shadow model to simulate the behavior of the reconstruction attacks to obtain the reconstruction loss which denotes the quality of the reconstructed images. Thereafter, we maximize the reconstruction loss to generate the adversarial features by iteratively adding the adversarial latent noise to features along the direction of the gradient (loss w.r.t. the targeted feature). Moreover, to ensure face recognition accuracy, the magnitude of adversarial latent noise would be constrained during the optimization.

Our main contributions are summarized as follows:
\begin{itemize}
\vspace{-2mm}
\item We propose a novel facial privacy-preserving method (namely AdvFace), which can generate privacy-preserving adversarial features against unknown reconstruction attacks while maintaining face recognition accuracy. Moreover, AdvFace requires no changes to the deployed face recognition model and thus can be integrated as a plug-in privacy-enhancing module into face recognition systems. 

\vspace{-2mm}
\item We unveil the rationale of the reconstruction attack and build a shadow model to simulate the behavior of the reconstruction attacks and generate adversarial features, which can disrupt the mapping from features to facial images by maximizing the reconstruction loss of the shadow model. 

\vspace{-2mm}
\item Extensive experimental results demonstrate that our proposed AdvFace outperforms the state-of-the-art facial privacy-preserving methods in terms of superior privacy protection performance while only incurring negligible face recognition accuracy loss. Moreover, the transferability of AdvFace is validated. That is, it can effectively resist different reconstruction networks. 
\end{itemize}

\vspace{-3mm}
\section{Related work}
\vspace{-2mm}
This section provides an overview of related works on face reconstruction attacks and face privacy protection. 

\vspace{-1mm}
\subsection{Face Reconstruction Attacks}
\vspace{-2mm}
In earlier works, Mignon et al.~\cite{mignon2013reconstructing} used the radial basis function regression to reconstruct faces from their features. Mohanty et al.~\cite{mohanty2007scores} used the inverse of the affine transformation model, which simulates the face recognition system, to reconstruct facial images. However, regression and affine transformation-based methods are no longer applicable when facial features are extracted by complex deep neural networks. Fredrikson et al.~\cite{fredrikson2015model} performed the face reconstruction by solving an optimization problem, which aims to generate the reconstructed images that minimize the distance between the targeted features and features from the reconstructed images. Similarly, Razzhigaev et al.~\cite{razzhigaev2020black} followed the same optimization objective and transformed the problem into the linear space of 2D Gaussian functions for higher efficiency. However, these optimization-based reconstruction attacks incur large computation costs even for reconstructing only one facial image. Hence, some recent works~\cite{cole2017synthesizing,dosovitskiy2016inverting, zhmoginov2016inverting, mai2018reconstruction,he2019model} used the reconstruction network trained by a large number of (image, feature) pairs to map the features back to the facial images. In this paper, given their powerful attacking performance with moderate attacking costs, we choose the reconstruction network-based attacks as our defense target.

\subsection{Face Privacy Protection}
\vspace{-2mm}
Several face privacy protection methods have recently been proposed, which can be divided into four categories. The encryption-based methods, such as homomorphic encryption~\cite{gentry2011implementing}, matrix encryption~\cite{kou2021efficient}, functional encryption~\cite{abdalla2015simple}, and randomized CNN with user-specific keys~\cite{mai2020secureface}, encrypted facial images or features, and then performed face recognition in the encrypted space to protect face privacy. However, the high computational overhead of exchanging and processing data is not suitable for face recognition networks that already consume a lot of computational resources. In~\cite{chamikara2020privacy, Mao2018APD}, differential privacy-based methods were incorporated to add carefully crafted noises to features or images to prevent leaking information that can distinguish faces. However, these methods suffer from a significant accuracy drop in face recognition tasks. The adversarial training-based methods~\cite{li2021deepobfuscator, xiao2020adversarial} strengthened face privacy by directly generating a facial feature that could resist reconstruction attacks through the adversarial training between the reconstruction network and the feature extractor. However, researchers in~\cite{li2021deepobfuscator} revealed that the accuracy of face recognition decreased significantly (from 99.97\% to 30.38\%) when dealing with the driver identity recognition task. Researchers in~\cite{ji2022privacy, mi2022duetface} proposed face privacy protection methods based on the frequency domain segmentation, which transforms the facial images into the frequency features and removes parts of the features that are important for image reconstruction but secondary to face recognition to thwart the reconstruction attacks while ensuring the accuracy of face recognition. However, ~\cite{ji2022privacy} struggled with the tradeoff between accuracy and privacy protection and our experimental results demonstrate that \cite{mi2022duetface} cannot resist powerful reconstruction attacks. In addition, the frequency domain-based methods require retraining and redeploying the face recognition network leading to significantly increased costs. In summary, the aforementioned methods are either incapable of defending against reconstruction attacks while maintaining recognition accuracy or incurring large computation overhead or redeployment costs.


\vspace{-1mm}
\section{Preliminary}
\vspace{-2mm}
In this section, we first introduce a typical architecture of face recognition systems and then present a realistic threat model considered in this paper.  

\vspace{-1mm}
\subsection{Face Recognition Systems}
\vspace{-2mm}

\begin{figure}[!t]
\centering
\includegraphics[width=0.95\columnwidth]{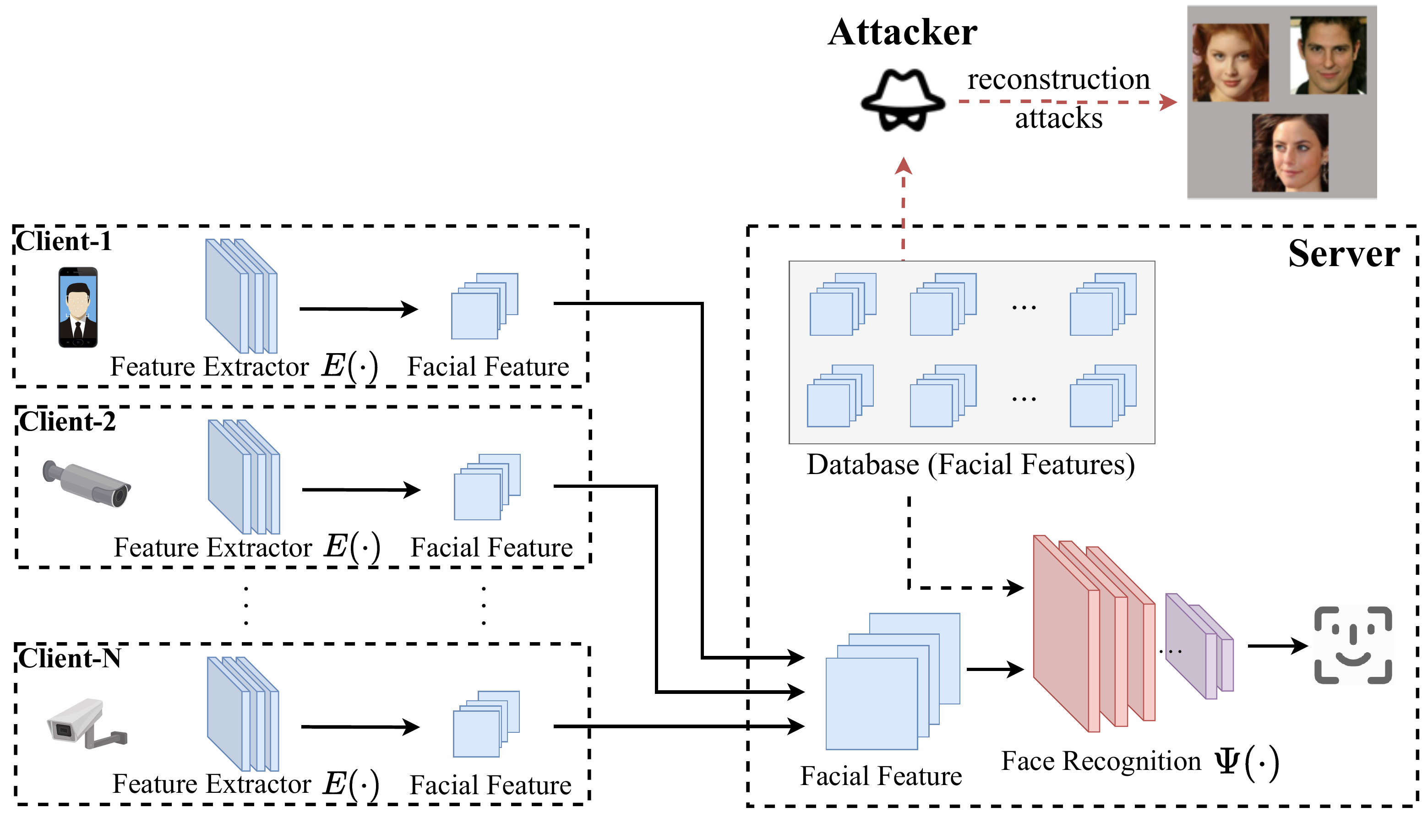}
\caption{The typical architecture of face recognition systems.}
\label{fig:tradition framework}
\vspace{-3mm}
\end{figure}

For face protection, existing face recognition systems usually use the client-server architecture~\cite{ko2018edge, eshratifar2019jointdnn}. As shown in Fig.~\ref{fig:tradition framework}, the face recognition network is partitioned and deployed as two sequential modules, i.e., the feature extractor $E(\cdot)$ (the front layers of the network) on the client side and the remaining layers $\Psi(\cdot)$ on the server side. In particular, each client uses $E(\cdot)$ to extract features from facial images and submits them to the server, where $\Psi(\cdot)$ is employed to recognize the identities of the received facial features.

Rather than original facial images, the server stores facial features that do not visually disclose facial information. However, as previously mentioned, once stolen from the server's database, the facial features can still be exploited to reconstruct facial images via the reconstruction attack. Therefore, effective face privacy-preserving methods that can protect the original facial images from being reconstructed from the facial features, are highly desired.

\begin{figure*}[!t]
\centering
\includegraphics[width=2\columnwidth]{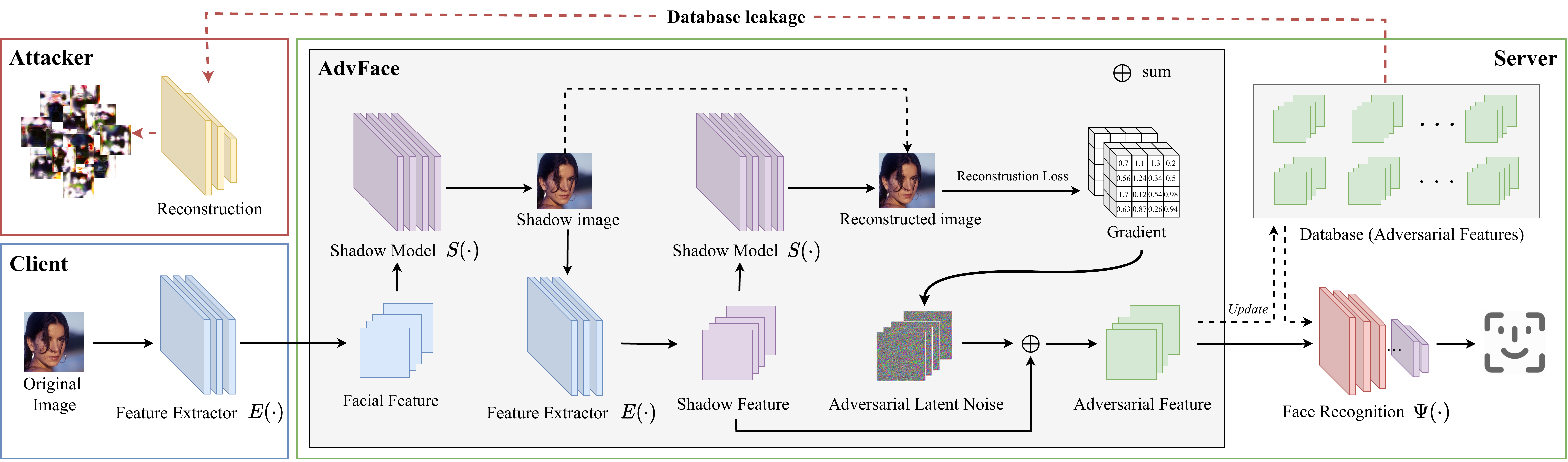}
\vspace{-3mm}
\caption{The pipeline of AdvFace in the deployed face recognition system. In the client, the original image is converted to a facial feature which will be uploaded to the server. In the server, the facial feature is reconstructed to a shadow image by the shadow model built by the server, and the shadow feature is extracted and further converted to the reconstructed image. Finally, the generation of the adversarial feature depends on the loss of the shadow image and the reconstructed image. The server stores adversarial features rather than original facial features uploaded from the client against reconstruction attacks.}
\label{fig:pipeline}
\vspace{-5mm}
\end{figure*}

\vspace{-1mm}
\subsection{Threat Model}\label{sec:Threat Model}
\vspace{-2mm}
In this paper, we consider that the server in the face recognition system is trusted. However, there may exist external attackers, who tend to steal the facial features from the server's database and launch reconstruction attacks to obtain clients' original facial images. 




\noindent\textbf{Attacker’s Knowledge}:
We consider that the attacker is powerful, and it has the following knowledge:
\begin{itemize}
    \vspace{-2mm}
    \item Facial features: The attacker can obtain the facial features stored in the server's database. 
    \vspace{-2mm}
    \item The Feature Extractor: The attacker can access the clients' black-box feature extractor of the face recognition model. Note that this can be easily achieved by purchasing a client device from the face recognition service provider.
\end{itemize}
\vspace{-2mm}

This powerful attacker also makes it difficult for us to design an effective facial privacy-preserving approach.

\noindent\textbf{Attacker’s Strategy}:
Following existing reconstruction attacks \cite{cole2017synthesizing, dosovitskiy2016inverting, zhmoginov2016inverting, mai2018reconstruction,he2019model}, we consider that the attacker tends to reconstruct facial images from facial features by building a reconstruction network, denoted by $R(\cdot)$. The attacker can train $R(\cdot)$ by minimizing the reconstruction loss function $\mathcal{L}_{R}$, which is defined as the $L_1$-norm distance between the original and reconstructed images. Formally, we have:
\vspace{-1mm}
\begin{equation}
\mathcal{L}_{R}(Z, X)=\sum_{i=1}^N \left\|x_i-R\left(z_i\right)\right\|_{1}, 
\label{eq:reconstruction}
\vspace{-1mm}
\end{equation}
where $x_i$ denotes a facial image in a public face dataset $X = \{x_1,\ldots ,x_N\}$ and $Z = \{z_1,\ldots, z_N\}$ (where $z_i = E(x_i)$) represents the corresponding features extracted by the feature extractor. Here, $N$ represents the total number of facial images in the public dataset.

With the trained reconstruction network $R(\cdot)$, the attacker can easily reconstruct facial images from the facial features via only one step of forward propagation on $R(\cdot)$. 
\vspace{-1mm}
\section{Adversarial Features Based Face Protection}
\vspace{-2mm}

In this section, we first provide an overview of AdvFace and then elaborate on the design of the shadow model design and the adversarial features generation. Finally, we present the application scenarios of AdvFace. 

\vspace{-1mm}
\subsection{Overview of AdvFace}
\vspace{-2mm}
The key to the success of reconstruction attacks is that they can learn the mapping from features to images by optimizing Eq.~\eqref{eq:reconstruction}. Although attackers may use different reconstruction network structures, similar mappings could be learned given the image-feature pairs extracted from the same feature extractor. Therefore, our basic idea is to learn the mapping function by building a shadow model of the reconstruction network and then disrupt the mapping from features to facial images along the opposite direction of training the reconstruction network. To this end, we craft the adversarial features with the adversarial latent noise by solving a constrained optimization problem, which aims to maximize the difference between the original images and the reconstructed images from the adversarial features through the shadow model.


Fig.~\ref{fig:pipeline} shows the workflow of AdvFace, which takes the original facial features uploaded from a client as the input and outputs the corresponding adversarial features that will be stored in the server's database for future online face recognition. It is worth noting that the original facial features would not be stored in the database. Thus, when the database is breached, only these adversarial features would be leaked, which could prevent the attacker from reconstructing the facial images.

Specifically, AdvFace consists of the following phases: 1) converting the original facial features uploaded from the client to shadow images with the shadow model; 2) extracting shadow features from shadow images with the feature extractor; 3) reverting the shadow features to the reconstructed images by the same shadow model, and calculating the gradients of $\mathcal{L}_S$ w.r.t. shadow features, which indicates the direction where $\mathcal{L}_R$ increases most quickly; 4) generating adversarial latent noise by gradient ascent; 5) perturbing the shadow features with the adversarial latent noise to generate the adversarial features, which will be stored in the database of the server for future face recognition.

\vspace{-1mm}
\subsection{Shadow Model Building}
\vspace{-2mm}
Despite being exploited by attackers to reconstruct facial images, the visual information contained in facial features is essential for face recognition. This, however, has been largely neglected in existing obfuscation-based privacy protection methods, which usually choose to distort facial features indiscriminately. Consequently, these methods sacrifice face recognition accuracy for privacy protection. To strike a desirable balance between privacy protection and face recognition, we need to craft adversarial features that maximize the reconstruction loss while minimizing the face recognition accuracy loss. Specifically, for a well-trained face recognition network, one feasible way of mitigating recognition accuracy loss is to maximize the reconstruction loss while enforcing a constraint on the magnitude of the disturbance on the features.


In particular, AdvFace aims to find an $L_p$-norm bounded noise $\delta$ to distort features such that the reconstruction loss $\mathcal{L}_{R}$ is maximized, which is formulated as the following constrained optimization problem:

\vspace{-5mm}
\begin{equation}
\mathop{\arg\max}\limits_{\delta} \|R(z+\delta) - x\|_{1}, \quad \text { s.t. }\|\delta\|_p < \xi,
\label{eq:goal}
\vspace{-2mm}
\end{equation}
where $x$ is an original facial image, $z$ represents the facial features extracted from $x$, $\delta$ denotes the adversarial latent noise, and $\xi$ represents the noise bound. Intuitively, the optimization problem~\eqref{eq:goal} can be easily solved by adding noises along the direction of the gradient of $\mathcal{L}_{R}$.

However, it is rather challenging or even impossible to directly solve~\eqref{eq:goal} as the reconstruction network employed by the attacker is unknown beforehand. Since different reconstruction networks learn a similar mapping from facial features to images, given the image-feature pairs extracted by the same feature extractor, the solution to problem~\eqref{eq:goal} is actually reconstruction model-agnostic. Hence, we advocate building a powerful shadow model $S(\cdot)$ at the server, which can be any reconstruction network, to learn the corresponding mapping and compute the reconstruction loss (then the corresponding gradients). Similar to~\eqref{eq:reconstruction}, we can train the shadow model $S(\cdot)$ on a public face dataset by minimizing the following loss function:
\vspace{-2mm}
\begin{equation}
\vspace{-1mm}
\mathcal{L}_{S}(Z, X)=\sum_{i=1}^N \left\|x_{i}-S\left(z_{i}\right)\right\|_{1}.
\label{eq:reconstruction_shadow}
\end{equation}

However, without accessing the original facial images, the server can not derive the reconstruction loss regarding the stored facial features. To address this issue, as shown in Fig.~\ref{fig:pipeline}, we first convert the features submitted by the client to shadow images through the shadow model and extract shadow features using the feature extractor. Then, we can compute the reconstruction loss and gradient based on the shadow images and images reconstructed from shadow features. The above process can be formally represented as
\vspace{-1mm}
\begin{equation}
\tilde{x} = S(z), \tilde{z} = E(\tilde{x}),
\vspace{-1mm}
\end{equation}
where $\tilde{x}$ is the shadow image reconstructed by the shadow model from the facial feature $z$ submitted by the client, and $\tilde{z}$ is the shadow feature extracted from the shadow image $\tilde{x}$ using the feature extractor. 

Then, we can calculate the gradient $grad(S,\tilde{z}+\delta,\tilde{x})$ of the reconstruction loss of the shadow model w.r.t. the perturbation $\delta$ as follows

\vspace{-1mm}
\begin{equation}
 grad(S,\tilde{z}+\delta,\tilde{x}) = \nabla_{\delta} \|S(\tilde{z}+\delta) - \tilde{x}\|_{1},
 \label{eq:directly solution}
 \vspace{-1mm}
\end{equation}
where $\tilde{z}+\delta$ denotes the adversarial feature with $\delta$ being initialized to zero. With noise added, the attacker can not recover the shadow image $\tilde{x}$ from the adversarial feature. Since $\tilde{x}$ is highly similar to the original image $x$, it is also hard for the attacker to reconstruct the original image. 

Furthermore, considering that the facial images used for training and those encountered after deploying the face recognition network can be quite different, we update the parameters of the batch normalization (BN) layer in the shadow model. Specifically, unlike a typical BN process, which normalizes inputs during the inference stage using the parameters learned from the training dataset, we compute the mean $\mu$ and variance $\sigma$ based on the transmitted testing feature batches to ensure that the noise added to each feature is more appropriate.

\subsection{Adversarial Features Generation}
To generate the adversarial features, we inject the adversarial latent noise $\delta$ into the shadow features under the guidance of $grad(S,\tilde{z}+\delta,\tilde{x})$ as in~\eqref{eq:directly solution}. Generally, we can use any gradient-based methods to generate adversarial features~\cite{madry2017towards, goodfellow2014explaining, dong2018boosting} to break the mapping from the features to the original facial images such that the face recognition system can defend against the reconstruction attack. In this paper, we choose the Project Gradient Descent (PGD)~\cite{madry2017towards}, which iteratively adds noises along the gradient direction while restricting the perturbation range in each iteration. Specifically, the generation of adversarial features can be formulated as: 

\vspace{-5mm}
 \begin{equation}
 \begin{aligned}
z_{t+1} =  z_{t} + \alpha & \cdot sign(grad(S,z_{t},\tilde{x})), \; z_{0} = \tilde{z}, \\ & \text{ s.t. }\|z_{t+1}-z_{t}\|<\varepsilon,
\end{aligned}
\label{eq:round perturb}
\end{equation}
where $\alpha$ controls the magnitude of noise and $\varepsilon$ restricts the noise level added in each iteration. Here, $sign\left(\cdot\right)$ is an element-wise function that outputs 1 for positive gradient values, -1 for negative gradient values, and 0 for 0. 

Starting from $\tilde{z}$, we update the adversarial feature by iteratively adding noises following~\eqref{eq:round perturb}. With the function $sign\left(\cdot\right)$, the noise added in each iteration is not exactly along the direction of the gradient but an approximate one. This helps alleviate the negative influences of some extreme samples, contributing to enhanced robustness of adversarial features.


\vspace{-1mm}
\subsection{Discussions of Application Scenarios}
\label{sub:discussion}
\vspace{-2mm}


We would like to strengthen that our proposed AdvFace can protect facial privacy without changing the face recognition networks. Thus, AdvFace can be easily integrated into deployed face recognition systems as a plug-in privacy-enhancing module. 

Moreover, AdvFace can work in both online and offline modes. Specifically, the server of a face recognition system can employ AdvFace to generate adversarial features from original facial features in real-time, i.e., online mode. However, with noises added, the adversarial features inevitably incur a slight decrease in face recognition accuracy. To tackle this issue, the server can use the adversarial features and labels stored in the database to retrain the face recognition network on the server side, which is the offline mode. The server-side face recognition network can learn sufficient information about the adversarial features through the offline mode, thereby improving face recognition accuracy. Note that the offline mode does not involve any changes to the adversarial features themselves. Thus, the privacy protection performance will not be compromised. Furthermore, the generated adversarial features can be packaged into privacy-preserving datasets for data sharing and reused for training other face recognition networks. 

\vspace{-2mm}
\section{Experimental Evaluation}
\vspace{-2mm}
In this section, we conduct extensive experiments on various datasets and models to evaluate the performance of AdvFace in terms of face recognition accuracy, attack resistance, and transferability.

\vspace{-1mm}
\subsection{Experimental Setup}\label{subsec:datasets}
\vspace{-1mm}
\subsubsection{Datasets}
\vspace{-1mm}
We use the following datasets in our experiments. 
\begin{itemize}
\vspace{-1mm}
\item \textit{CASIA-WebFace}~\cite{yi2014learning} contains 490K facial images from more than 10k different individuals.
\vspace{-1mm}
\item \textit{CelebA}~\cite{liu2015deep} contains 202K facial images from more than 10k celebrities.
\vspace{-1mm}
\item \textit{LFW}~\cite{huang2008labeled} contains 13K facial images from 5.7K different identities and 6K face pairs (i.e., two facial images) for evaluation.
\vspace{-1mm}
\item \textit{CFP-FP}~\cite{sengupta2016frontal} contains 7K images from 500 identities and 7K face pairs for evaluation.
\vspace{-1mm}
\item \textit{AgeDB-30}~\cite{moschoglou2017agedb} contains more than 12K images from 570 identities and 6K face pairs for evaluation.
\end{itemize}





Following the preprocessing operation adopted in prior works ~\cite{deng2019arcface, mai2020secureface}, we crop all facial images with the multi-task convolutional neural network (MTCNN)~\cite{zhang2016joint}, which can detect faces and facial landmarks in images. Besides, for each cropped image, we resize it to 160 $\times$ 160 for a fair comparison. Moreover, each pixel in RGB format (i.e., [0,255]) is normalized to [0,1] before being fed into the neural network.

\vspace{-3mm}
\subsubsection{Models and Implementation Details}\label{sec:Models and Implementation Details}
\vspace{-1mm}

\noindent\textbf{Face Recognition Model}:
We employ the FaceNet~\cite{schroff2015facenet} with a pre-trained Inception-ResNet-v1~\cite{he2016deep} as the backbone for face recognition. We select the first three convolutional layers of the backbone as the feature extractor $E(\cdot)$ deployed on the client-side, while the remaining layers are deployed on the server-side. We fine-tune the classifier of FaceNet for 50 epochs with the backbone frozen using the CASIA-WebFace dataset, and then fine-tune the entire network for 50 epochs. We use the Adam optimizer~\cite{kingma2014adam} with a scheduler, where the period and multiplicative factor of learning rate decay are set to 1 and 0.94. The entire model is trained with triplet loss~\cite{balntas2016learning} and cross-entropy loss.

\noindent\textbf{Face Reconstruction Model}: As summarized in the appendix, three types of reconstruction networks can be employed by the attacker. Specifically, the URec is built based on the U-net~\cite{ronneberger2015u} architecture and the ResRec is implemented with the ResNet~\cite{he2016deep} architecture. The TransRec is an exactly mirrored reconstruction network by performing a layer-to-layer reversion. All reconstruction networks are trained on the CelebA dataset. 

\noindent\textbf{AdvFace Model}: To implement AdvFace, we also train three types of shadow models $S(\cdot)$, including URec, ResRec, and TransRec on the CASIA-WebFace dataset. We adopt the Adam optimizer with a learning rate of 1e-4.  In the PGD process, we implement 40 iterations with $\alpha = 0.2$. Unless otherwise specified, the noise bound is $\varepsilon = 0.2$. 



\vspace{-2mm}
\subsubsection{Baseline Defense Methods}
\vspace{-2mm}
We compare AdvFace with the following three widely used face privacy protection methods.

\noindent\textbf{Random Perturbation}: This method iteratively adds randomly generated noises to the original facial features. To ensure a fair comparison, the total number of iterations is 40 and the noise bound in each iteration is 0.2. 

\noindent\textbf{Differential Privacy (DP)}~\cite{li2021deepobfuscator}: This method adds noises generated from the Laplace mechanism to the original facial features. The privacy budget is set to 1.0 to ensure the same noise bound as our method.

\noindent\textbf{DuetFace}~\cite{mi2022duetface}: This is the latest face privacy protection method based on frequency domain segmentation. It splits the frequency channels into two parts according to their importance for visualization and mainly use the non-crucial part for transmission and face recognition. 

\vspace{-2mm}
\subsubsection{Evaluation Metrics}
\vspace{-2mm}
To evaluate the performance of privacy protection methods against reconstruction attacks, we use SSIM \cite{wang2004image}, PSNR, and MSE to measure the quality of reconstructed images. Specifically, a larger MSE or a smaller SSIM and PSNR, indicates a lower similarity between the reconstructed image and the original facial image, which implies a stronger defense. For replay attacks which use the reconstructed images to cheat the face recognition system, we use the success rate of replay attacks (SRRA) to measure the performance of the protected features. A lower SRRA indicates that the protected facial feature has a stronger defense ability against the replay attack. Moreover, we characterize the utility of face recognition using the accuracy (ACC) of identifying whether two face features (from face pairs in LFW, CFP-FP, and AgeDB-30) belong to one person. 


\begin{table*}[]
\centering
\caption{Performance comparison results among privacy protection methods in terms of SSIM, PSNR, MSE, and SRRA.}
\vspace{-3mm}
\resizebox{1.9\columnwidth}{!}{
\begin{tabular}{lcccccccccccccc}
\toprule
            & \multicolumn{4}{c}{LFW} &  & \multicolumn{4}{c}{CFP-FP} &  & \multicolumn{4}{c}{AgeDB-30} \\ \cline{2-5} \cline{7-10} \cline{12-15} 
Methods     & SSIM$\downarrow$  & PSNR$\downarrow$  & MSE$\uparrow$   & SRRA$\downarrow$    &  & SSIM$\downarrow$  & PSNR$\downarrow$   & MSE$\uparrow$    & SRRA$\downarrow$     &  & SSIM$\downarrow$   & PSNR$\downarrow$    & MSE$\uparrow$    & SRRA$\downarrow$     \\ \hline
Unprotected & 0.93 & 27.87 & 0.002                   & 97.40\%  & & 0.83 & 22.89 & 0.006                   & 89.71\%  & & 0.87 & 23.96 & 0.005                   & 84.53\% \\
Random      & 0.90 & 22.81 & 0.005                   & 94.73\%  & & 0.79 & 20.73 & 0.009                   & 87.26\%  & & 0.86 & 21.68 & 0.007                   & 77.47\% \\
DP          & 0.90 & 23.12 & 0.005                   & 93.97\%  & & 0.79 & 20.89 & 0.009                   & 84.94\%  & & 0.86 & 21.86 & 0.007                   & 78.07\% \\
DuetFace    & 0.85 & 20.92 & 0.009                   & 95.17\%  & & 0.66 & 14.38 & 0.043                   & 70.23\%  & & 0.76 & 14.65 & 0.040                   & 87.27\% \\ 
\rowcolor{mygray}
Ours        & 0.28  & 6.97  & 0.206 & 4.03\%  &  & 0.23  & 5.98   & 0.261  & 18.43\%  &  & 0.24   & 5.85    & 0.269  & 16.67\%  \\ 
\bottomrule
\end{tabular}}
\label{table:ssim psnr mse}
\end{table*}


\vspace{-1mm}
\subsection{Trade-off between Privacy and Utility}\label{sec:trade-off}
\vspace{-2mm}
We first evaluate the effectiveness of AdvFace in terms of the trade-off between face privacy protection and face recognition accuracy. To this end, we perturb the features with different noise bounds, i.e., $\varepsilon$ varies from 0.00 to 0.30 with a step size of 0.05, where $\varepsilon = 0.00$ indicates no protection for the shadow features. We conduct experiments on three datasets, i.e., LFW, AgeDB-30, and CFP-FP, where the ResRec is used as the reconstruction network and the shadow model in AdvFace. 

\begin{figure}[!t]
\centering
\includegraphics[width=0.95\columnwidth]{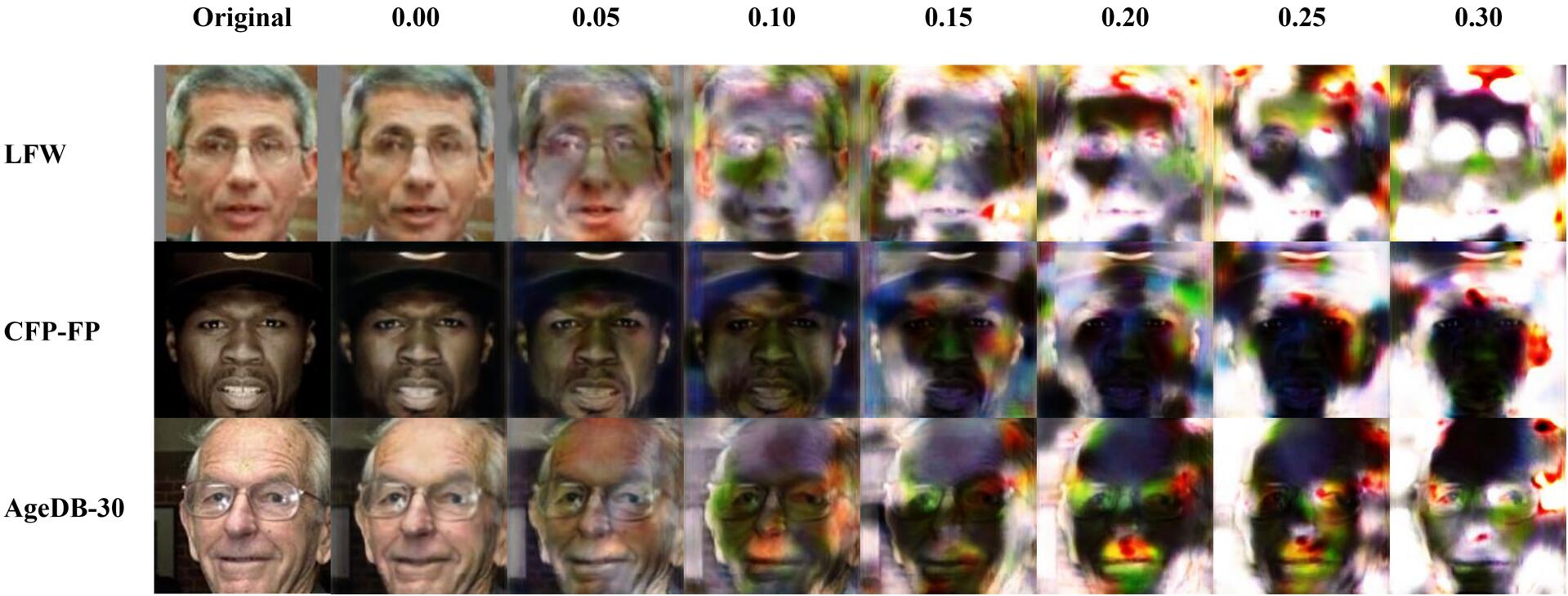}
\vspace{-2mm}
\caption{Reconstructed images from adversarial features with different noise bounds on datasets LFW, CFP-FP, and AgeDB-30.}
\label{fig: trade-off reconstruted img}
\vspace{-0.2cm}
\end{figure}

\begin{figure}[!t]
\centering
\includegraphics[width=0.7\columnwidth]{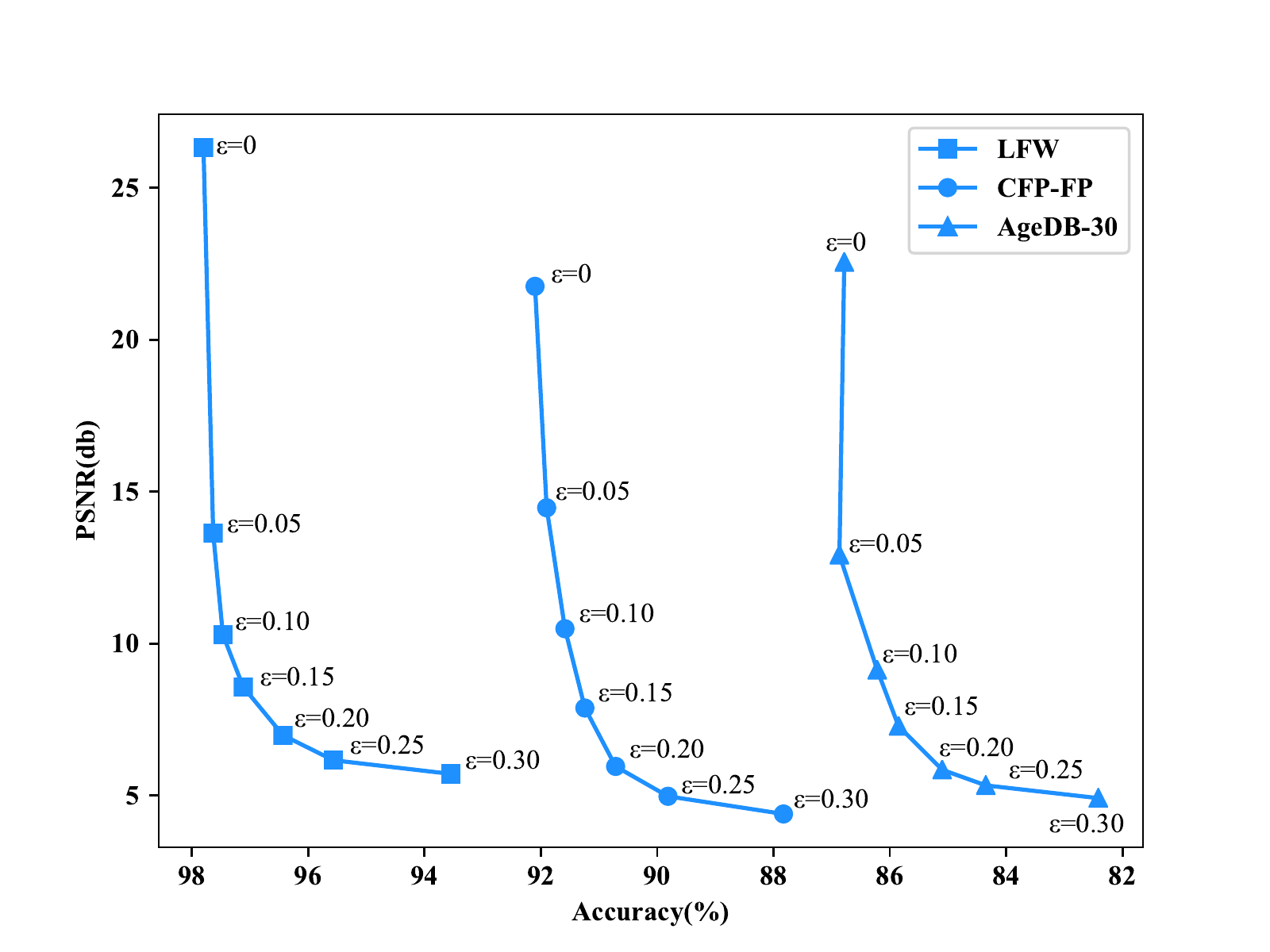}
\vspace{-2mm}
\caption{Performance of AdvFace in terms of the trade-off between PSNR and Accuracy with different noise bounds.}
\label{fig: trade-off PSNR-Acc}
\vspace{-6mm}
\end{figure}

Fig.~\ref{fig: trade-off reconstruted img} shows that the strength of privacy protection is proportional to $\varepsilon$. That is, a larger $\varepsilon$ always provides a stronger privacy protection. Moreover, we can see that face privacy can be well protected when $\varepsilon \geq 0.2$.  Fig.~\ref{fig: trade-off PSNR-Acc} shows the accuracy and PSNR values under different $\varepsilon$. We can see that the accuracy decreases as $\varepsilon$ increases, which is opposite to the changing trend of privacy protection (i.e., decreasing PSNR means improving privacy protection). The bottom left of Fig.~\ref{fig: trade-off PSNR-Acc} presents a good performance on both face privacy protection and face recognition accuracy. It can be seen that the point of $\varepsilon = 0.2$ is closest to the bottom left, and thus we set $\varepsilon$ as $0.2$ for the following evaluation.

\vspace{-1mm}
\subsection{Defense against Privacy Attacks}\label{sec:defend}
\vspace{-2mm}
We now compare AdvFace with baselines to evaluate its defense performance against privacy attacks (including reconstruction attacks and replay attacks) on the datasets of LFW, CFP-FP, and AgeDB-30. Both the reconstruction network and the shadow model adopt the ResRec architecture.



\begin{figure}[!t]
\centering
\includegraphics[width=0.8\columnwidth]{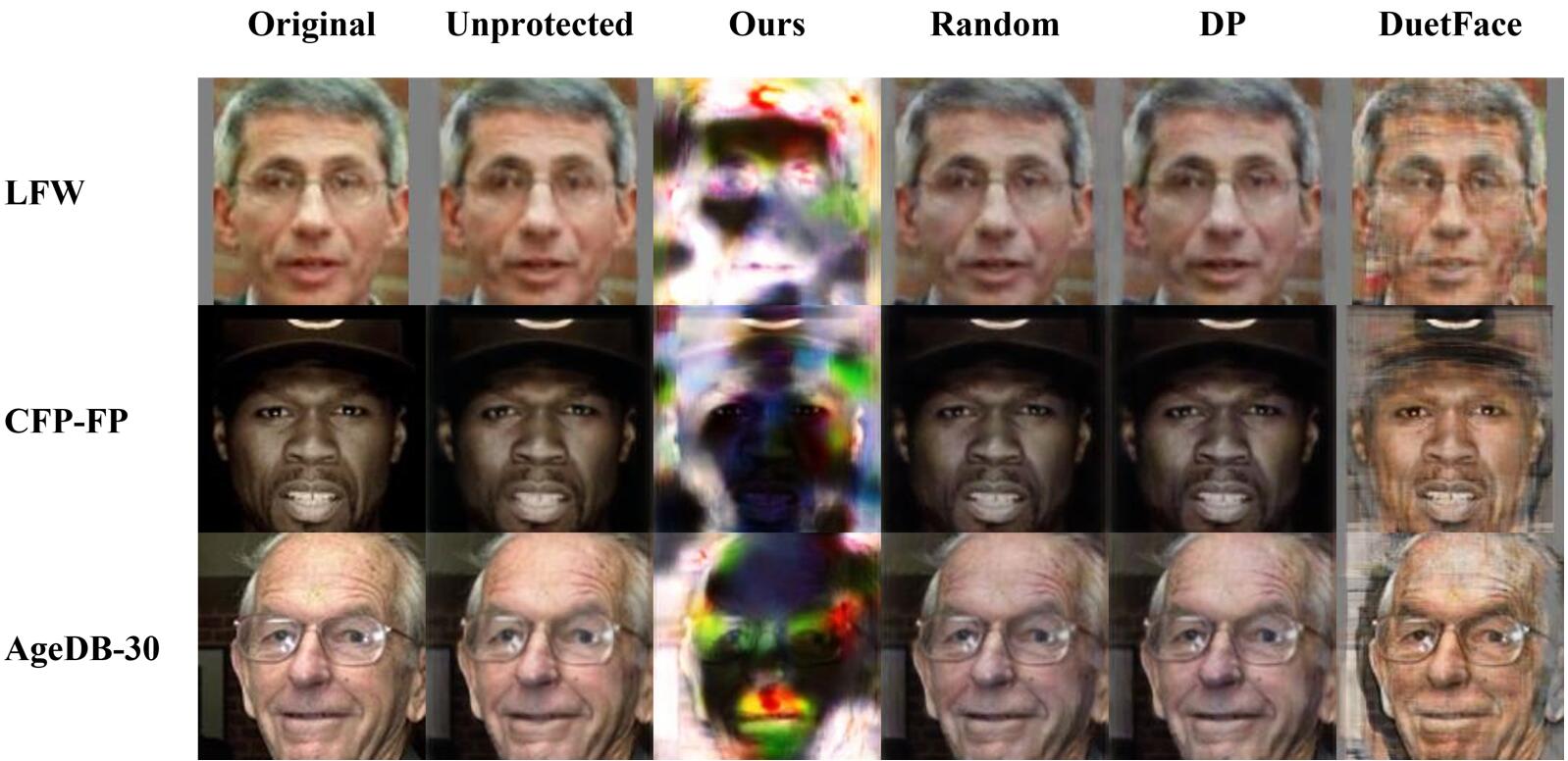}
\vspace{-2mm}
\caption{Reconstructed images from facial features generated by different privacy protection methods.}
\label{fig: Visualization of reconstruction attacks for different methods}
\vspace{-6mm}
\end{figure}

\noindent\textbf{Defense against Reconstruction Attacks}: Fig.~\ref{fig: Visualization of reconstruction attacks for different methods} shows the reconstructed images from facial features protected by different methods. As shown in the third column, the reconstructed images from the generated adversarial features by our proposed AdvFace are hard to distinguish, while those protected by other methods (columns 4-6) undergo much information leakage about the original images, which can identify the person. Tab.~\ref{table:ssim psnr mse} summarizes the average values of SSIM, PSNR, and MSE of the reconstructed images and original images. We can see that our method always has a lower SSIM/PSNR and a higher MSE, which demonstrates that our AdvFace outperforms other protection methods on the defense performance against the reconstruction attacks.

\noindent\textbf{Defense against Replay Attacks}: These attacks input the reconstructed facial image to the face recognition system for malicious face authentication, where we use the SRRA to measure the defense effectiveness. Tab.~\ref{table:ssim psnr mse} shows the outstanding performance of AdvFace in preventing attackers from launching replay attacks. Specifically, the value of SRRA significantly decreases (from 97.40\% to 4.03\%) after using AdvFace for privacy protection. In contrast, other protection methods fail to resist the replay attacks, with the SRRA being 94.73\%, 93.97\%, and 95.17\% for Random Perturbation, DP, and DuetFace, respectively. 

\begin{table}[!t]
\centering
\caption{The performance of privacy protection methods in terms of face recognition accuracy (note that Ours (online) represents that AdvFace is employed as a plug-and-play privacy-enhancing module and Ours (offline) indicates that we use the adversarial features to retrain the downstream face recognition network to further improve the accuracy).}
\vspace{-2mm}
\scalebox{0.8}[0.8]{
\begin{tabular}{lccc}
\toprule
Mehods        & LFW     & CFP-FP  & AgeDB-30 \\ \hline
Unprotected   & 98.13\% & 93.16\% & 87.57\%  \\
Random        & 97.20\% & 91.67\% & 86.60\%  \\
DP            & 96.27\% & 90.84\% & 85.12\%  \\
DuetFace      & 98.02\% & 84.37\% & 87.10\%  \\ 
\rowcolor{mygray}
Ours(online)  & 96.43\% & 90.59\% & 85.10\%  \\
\rowcolor{mygray}
Ours(offline) & 97.78\% & 92.04\% & 86.35\%  \\ 
\bottomrule
\end{tabular}}
\label{table: Face recognition accuracy after privacy protection}
\end{table}

\vspace{-1mm}
\subsection{Accuracy of Face Recognition}\label{sec:acc}
\vspace{-2mm}
In this subsection, we compare AdvFace with baseline methods to evaluate its performance in face recognition accuracy. As mentioned in Sec.~\ref{sub:discussion}, AdvFace can work in the online mode or offline mode according to whether the face recognition network is retrained or not. Note that, unless specifically marked as offline mode, the AdvFace works in the online mode. 

As shown in Tab.~\ref{table: Face recognition accuracy after privacy protection}, the online AdvFace integrated into the unprotected methods causes a small drop in accuracy (i.e., 1.7\%, 2.57\%, and 2.47\% lower than the unprotected method on the three datasets). However, we would like to clarify that such a slight decrease in accuracy is acceptable given the outstanding privacy protection performance of AdvFace. Furthermore, when AdvFace works in the offline mode, it achieves comparable accuracy with other protection methods. For instance, the accuracy of offline AdvFace is only 0.35\%, 1.12\%, and 1.22\% lower than the unprotected benchmark.



\begin{figure}[!t]
\centering
\includegraphics[width=0.85\columnwidth]{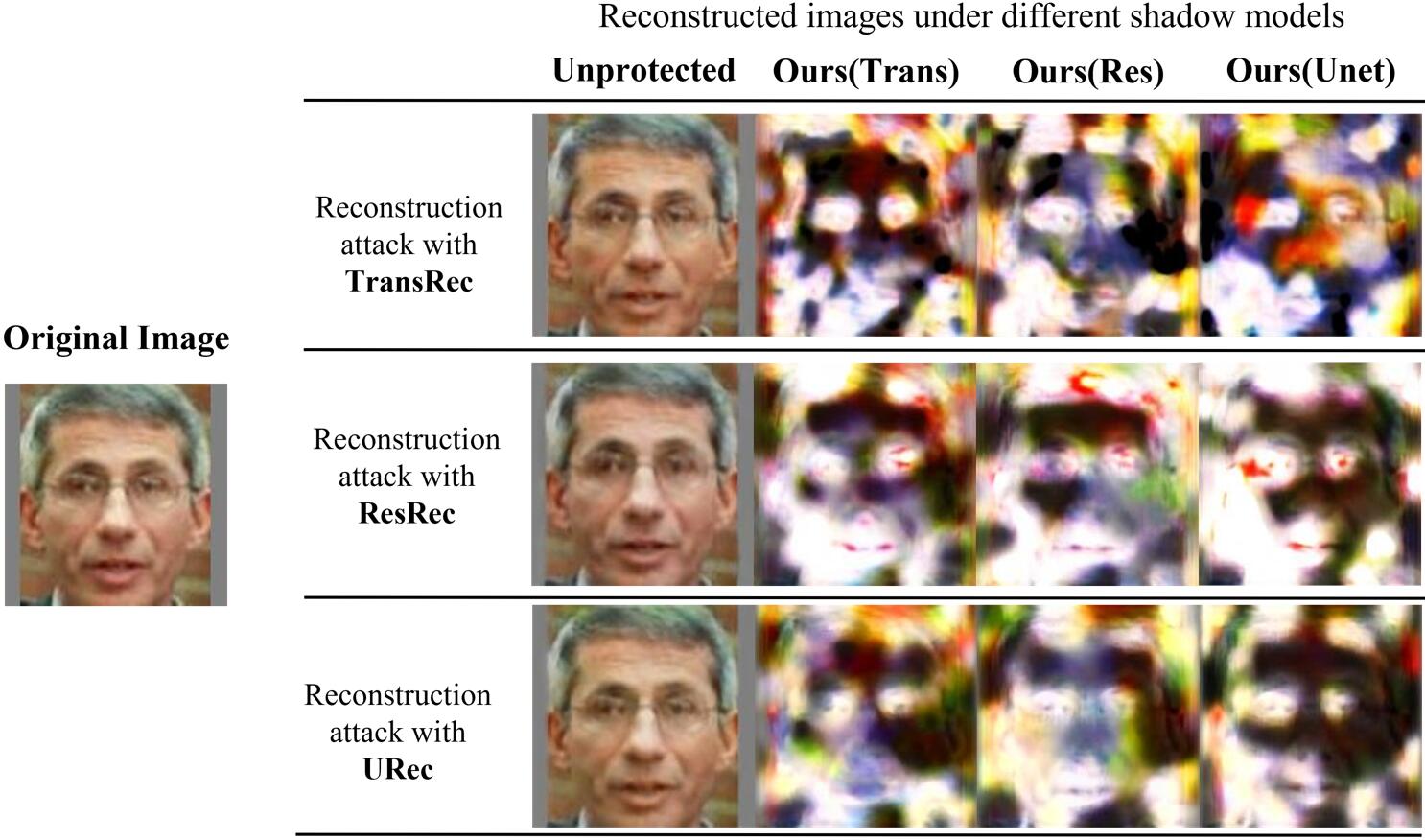}
\vspace{-2mm}
\caption{Transferability of AdvFace on defending against reconstruction attacks. Ours(Trans), Ours(Res), and Ours(Unet) represent that we adopt the transpose, resnet, and unet network as the shadow model, respectively. TransRec, ResRec, and URec represent that the transpose, resnet, and unet network is employed as the reconstruction network, respectively.}
\label{fig: scalability img}
\vspace{-4mm}
\end{figure}

\subsection{Transferability of AdvFace}\label{sec:Transferability}
\vspace{-2mm}
This subsection investigates the impact of the shadow model structure on AdvFace and the transferability of adversarial features generated by AdvFace. To this end, we first evaluate the performance of AdvFace on face recognition accuracy with different shadow model structures. Then, we evaluate the performance of AdvFace with different shadow model structures regarding the defense effectiveness against different reconstruction networks.



\begin{table}[!t]
\centering
\caption{The performance of AdvFace in the face recognition accuracy with different shadow models.}
\vspace{-3mm}
\scalebox{0.85}[0.85]{
\begin{tabular}{lccc}
\toprule
Protect Method & LFW & CFP-FP & AgeDB-30 \\ \hline
Ours(Trans)         & 96.53\%      & 90.77\%         & 85.13\%           \\
Ours(Res)            & 96.43\%      & 90.59\%         & 85.10\%           \\
Ours(Unet)              & 96.42\%      & 90.13\%         & 84.58\%           \\ 
\bottomrule
\end{tabular}}
\label{table: Acc of scalabitity}
\end{table}

Tab.~\ref{table: Acc of scalabitity} shows that similar face recognition accuracies are achieved by AdvFace with different shadow models, implying that the shadow model structure has only a slight influence on the performance of AdvFace. Thus, AdvFace is easy to implement. 


In Fig.~\ref{fig: scalability img}, we show the facial images reconstructed from the adversarial features by three different reconstruction networks. In Tab. \ref{table: SSIM and PSNR in scalability}, we quantitatively describe the average quality of reconstructed images with SSIM and PSNR on three datasets.  We can see the defense effectiveness of AdvFace is maintained when encountering different attack networks, which validates the transferability of the adversarial features generated by AdvFace. Specifically, the adversarial features generated by AdvFace (based on any shadow model) can defend against different reconstruction attacks.

\begin{table}[!t]
\centering
\caption{The performance of AdvFace in terms of SSIM and PSNR under different reconstruction networks.}
\vspace{-3mm}
\resizebox{0.95\columnwidth}{!}{
\begin{tabular}{cclccc}
\toprule
\textbf{Metric}                            & \textbf{Protect Method}           & \multicolumn{1}{c}{\textbf{Attack Method}} & \textbf{LFW} & \textbf{CFP-FP} & \textbf{AgeDB-30} \\ \hline
\multicolumn{1}{c|}{\multirow{9}{*}{SSIM}} & \multirow{3}{*}{Ours(transpose)} & TransRec                                    & 0.20         & 0.16            & 0.19              \\
\multicolumn{1}{c|}{}                      &                                  & ResRec                                      & 0.27         & 0.20            & 0.23              \\
\multicolumn{1}{c|}{}                      &                                  & UReC                                        & 0.26         & 0.20            & 0.22              \\ \cline{2-6} 
\multicolumn{1}{c|}{}                      & \multirow{3}{*}{Ours(resnet)}    & TransRec                                    & 0.23         & 0.19            & 0.21              \\
\multicolumn{1}{c|}{}                      &                                  & ResRec                                      & 0.28         & 0.23            & 0.24              \\
\multicolumn{1}{c|}{}                      &                                  & UReC                                        & 0.29         & 0.24            & 0.24              \\ \cline{2-6} 
\multicolumn{1}{c|}{}                      & \multirow{3}{*}{Ours(Unet)}      & TransRec                                    & 0.24         & 0.20            & 0.23              \\
\multicolumn{1}{c|}{}                      &                                  & ResRec                                      & 0.27         & 0.21            & 0.23              \\
\multicolumn{1}{c|}{}                      &                                  & UReC                                        & 0.28         & 0.23            & 0.23              \\ \hline
\multicolumn{1}{c|}{\multirow{9}{*}{PSNR}} & \multirow{3}{*}{Ours(transpose)} & TransRec                                    & 6.93         & 7.40            & 5.92              \\
\multicolumn{1}{c|}{}                      &                                  & ResRec                                      & 6.84         & 5.96            & 5.59              \\
\multicolumn{1}{c|}{}                      &                                  & UReC                                        & 7.33         & 6.81            & 6.14              \\ \cline{2-6} 
\multicolumn{1}{c|}{}                      & \multirow{3}{*}{Ours(resnet)}    & TransRec                                    & 6.90         & 7.01            & 5.92              \\
\multicolumn{1}{c|}{}                      &                                  & ResRec                                      & 6.97         & 5.98            & 5.85              \\
\multicolumn{1}{c|}{}                      &                                  & UReC                                        & 7.47         & 6.70            & 6.20              \\ \cline{2-6} 
\multicolumn{1}{c|}{}                      & \multirow{3}{*}{Ours(Unet)}      & TransRec                                    & 6.73         & 7.36            & 5.88              \\
\multicolumn{1}{c|}{}                      &                                  & ResRec                                      & 6.57         & 6.17            & 5.48              \\
\multicolumn{1}{c|}{}                      &                                  & UReC                                        & 7.01         & 6.94            & 5.95              \\ 
\bottomrule
\end{tabular}}
\label{table: SSIM and PSNR in scalability}
\vspace{-3mm}
\end{table}

\vspace{-2mm}
\section{Conclusions}
\vspace{-2mm}
In this work, we proposed an adversarial features-based face privacy protection (AdvFace) method to generate privacy-preserving adversarial features against the reconstruction attack while maintaining face recognition accuracy. Extensive experimental results show the superior performance of AdvFace in defending against reconstruction attacks compared to those state-of-the-art methods. At the same time, AdvFace can be easily integrated into deployed face recognition systems as a plug-in privacy-enhancing module. Moreover, the experiments also validate that AdvFace can achieve a desirable tradeoff between accuracy and utility and generate adversarial features with excellent transferability, promising its practicality and applicability.

\section*{Acknowledgments}
\vspace{-1mm}
This work was supported by Key R\&D Program of Zhejiang (Grant No. 2022C01018), National Natural Science Foundation of China (Grants No. 62122066, U20A20182, 61872274, 62102337) and National Key R\&D Program of China (Grant No. 2021ZD0112803).

{\small
\bibliographystyle{ieee_fullname}
\bibliography{egbib}
}

\end{document}